\documentclass{article}
\usepackage{iclr2027_conference,times}
\iclrfinalcopy

\usepackage[utf8]{inputenc} 
\usepackage[T1]{fontenc}    
\usepackage{hyperref}       
\usepackage{url}            
\usepackage{booktabs}       
\usepackage{amsfonts}       
\usepackage{nicefrac}       
\usepackage{microtype}      
\usepackage{xcolor}         
\usepackage{graphicx}
\usepackage{amsmath}
\usepackage{hyperref}
\usepackage[capitalise,nameinlink]{cleveref}
\usepackage{enumitem}
\usepackage{multirow}
\usepackage{stfloats}
\usepackage{subcaption}
\usepackage[normalem]{ulem}
\usepackage{xcolor}
\usepackage{wrapfig}
\usepackage{float} 

\usepackage{newunicodechar}
\newunicodechar{↑}{\ensuremath{\uparrow}}
\newunicodechar{↓}{\ensuremath{\downarrow}}
\newunicodechar{’}{\textquoteright}

\title{\textit{Back to the Future:} Reverse-Consequence Modeling for Long-Horizon Particle Dynamics}

\author{%
  Leyan Zhang, Nantian He, Jiajun Ma, Yanpeng Zhao, Yunbo Wang \\
  Shanghai Jiaotong University
}

\begin{document}

\maketitle
\lhead{}

\begin{abstract}
Accurate prediction of object trajectories is fundamental to mechanical understanding and robotic control. However, autoregressive models inherently suffer from cumulative errors, leading to severe drift in long-horizon rollouts. Observing that in many physical systems with dissipative or convergence-to-rest behavior, the final states are often more predictable than transient dynamics (\textit{e.g.}, a sliding object resting under friction or a pendulum stabilizing at equilibrium), we propose \textit{ReStep}, a cascaded approach that reverses the conventional forward-prediction paradigm. ReStep first employs a stochastic model to directly predict the final state, then leverages \textit{Anchored Diffusion} to generate goal-conditioned intermediate trajectories. By conditioning on both ``boundary states'', it effectively constrains dynamics and reduces uncertainty, producing physically plausible and temporally coherent transitions. We further integrate ReStep dynamics as the physical engine within a 3D Gaussian Splatting pipeline, enabling a unified physical-visual framework for intuitive inference from images.
Extensive experiments on diverse rigid and non-rigid collision datasets show that ReStep achieves superior long-term accuracy, substantially mitigating error accumulation via reverse-consequence modeling.
\end{abstract}

\vspace{-4pt}
\section{Introduction}
\vspace{-2pt}
Predicting the evolution of particle-based object dynamics over time is a fundamental challenge in physical simulation~\citep{battaglia2013simulation, dpi, pfaff2020learning, sanchez2018graph}, and is particularly critical for downstream tasks such as visual rendering, motion planning, and robotic manipulation~\citep{comphy, 2016learning, licontact, lindiffskill, marisp}.
Existing approaches typically follow an autoregressive paradigm~\citep{sgnn}, where future states are sequentially predicted based on previously generated frames.
While conceptually straightforward, these methods inherently suffer from \textbf{error accumulation}, such that small deviations at each timestep compound over time, causing trajectories to gradually veer off course and deviate from physically consistent dynamics. 
Therefore, they often fail to capture long-range dependencies or convergence behaviors common in real-world physical systems, such as frictional settling or contact stabilization.

\begin{figure*}[t]
\vspace{-10pt}
    \centering
    \includegraphics[width=\textwidth]{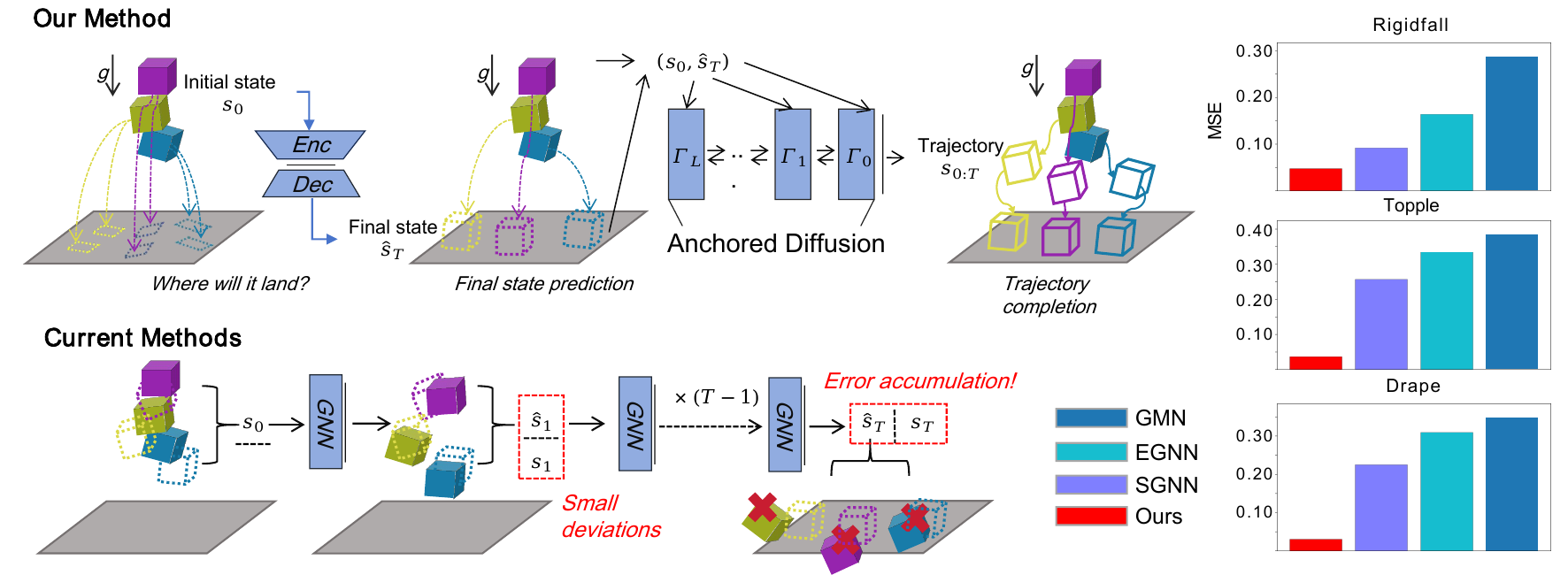}
    \vspace{-15pt}
    \caption{
    \textit{Overview of the proposed ReStep framework.} \textbf{Top:} A VAE-based steady-state predictor estimates plausible outcomes, which drives the Anchored Diffusion model to generate complete trajectories. \textbf{Bottom:} Traditional autoregressive models. Our approach mitigates this issue by conditioning trajectory generation on both initial and final constraints. The right-hand bar chart compares the Final State MSE of our method with that of autoregressive methods.
    }    
    \label{fig:pipeline}
\vspace{-5pt}
\end{figure*}

We observe a notable property in dissipative physical scenes that exhibit convergence behaviors: \textit{The final state, namely the steady state, is reasonably more predictable for neural networks to infer than the detailed transient trajectory leading to it}, under initial conditions and constraints.
For instance, when an object falls or collides, its eventual resting configuration is largely determined by global constraints such as gravity, mass distribution, and surface topology. This insight motivates a fundamental question:
\vspace{-4pt}
\begin{quote}
    \textit{What if we predict the outcome first, and then generate the trajectory backwards?}
\end{quote}
\vspace{-4pt}
Inspired by this, we propose ReStep, a two-stage approach that reframes the conventional forward-prediction paradigm as a backward-conditioned generation task. 
As shown in Figure \ref{fig:pipeline}, ReStep first employs a stochastic steady-state predictor based on VAE~\citep{vae} to estimate a plausible outcome directly from the initial conditions. 
Unlike previous particle dynamics prediction methods, such as SGNN~\citep{sgnn}, which do not account for the inherent uncertainty in physical interactions and can yield disentangled predictions over long horizons, ReStep explicitly models the distribution of plausible final states once the dynamic process has stabilized.

In the second stage, we introduce a \textit{Future-Anchored Diffusion Model} to generate full trajectories conditioned on both the initial and predicted final states, enabling temporally coherent transitions. 
While autoregressive models tend to accumulate errors over long horizons, diffusion models inherently mitigate this issue through iterative denoising, yielding stable long-term generation. Moreover, their strong generative and representational capabilities facilitate a more comprehensive understanding of complex physical dynamics.
The denoising network is designed to be fully convolutional along the particle dimension, enabling flexible processing of variable-sized particle sets. Therefore, Anchored Diffusion can scale seamlessly to a wide range of object and particle counts.

Beyond particle-space trajectory prediction, we note that simulation particles and their physical parameters are often unobservable in real-world settings. To bridge this gap, we integrate ReStep as the physical engine within a 3D Gaussian Splatting (3DGS) pipeline.
By coupling future-anchored dynamics with high-fidelity rendering, our approach enables direct inference from images, supporting tasks such as latent property estimation and physically consistent visual dynamics generation.


We conduct experiments on the \textit{FlexiFall} and \textit{Physion} benchmarks, covering both particle-based dynamics prediction and vision-grounded tasks for rigid and non-rigid objects. 
Our results show that ReStep improves long-horizon prediction accuracy, enables effective recovery of latent physical parameters from visual observations, and produces visually faithful scene dynamics by mitigating cumulative errors through reverse-consequence modeling.

In summary, our key contributions are as follows:
    \vspace{-4pt}
\begin{itemize}[leftmargin=*]
    \item We propose ReStep and reformulate long-horizon trajectory generation via reverse-consequence modeling, anchoring predictions on inferred future states to effectively eliminate cumulative drift.
    \vspace{-2pt}
    \item We develop a scalable VAE-based predictor that encodes physical interactions via GNNs and models the multimodal distribution over plausible future steady states.
    \vspace{-2pt}
    \item We introduce Anchored Diffusion, a future-conditioned generative model that synthesizes temporally coherent particle trajectories and improves long-horizon rollout accuracy.
    \vspace{-2pt}
     \item We incorporate ReStep as the dynamics module in 3D Gaussian Splatting, enabling intuitive inference of latent physical properties directly from raw images.
     \vspace{-4pt}
\end{itemize}

\vspace{-4pt}
\section{Related Work}
\vspace{-2pt}

\paragraph{GNN-based dynamics modeling.}
Graph neural networks (GNNs)~\citep{sperduti1997supervised,hamilton2017inductive, gilmer2017neural} have shown to be powerful for modeling particle-based physical systems~\citep{dpi, gmn, egnn, gns, sgnn, xu2024equivariant, liu2024segno}. Specifically, the graph is constructed by considering the pairwise distances among particles. Its evolution over time is derived through several iterations of message passing, where each node is updated by aggregating information from its neighbors. Some works~\citep{gmn,egnn,sgnn} leverage equivariant graph neural networks~\citep{han2022geometrically, xu2024equivariant, liu2024segno} to encode the inherent symmetries of 3D Euclidean space, achieving state-of-the-art performance in predicting physical dynamics. Despite strong performance in one-step transition prediction, these models suffer from error accumulation during trajectory rollouts.

\vspace{-8pt}
\paragraph{Diffusion models for conditional generation.}
Recent advances have introduced various techniques that enable diffusion models to perform conditional generation. Among them, classifier guidance~\citep{dhariwal2021diffusion} is a training-free method that uses a separately trained classifier to steer the sampling process toward the desired condition. Subsequently, classifier-free guidance~\citep{cfg} was introduced to remove reliance on an auxiliary classifier by jointly optimizing conditional and unconditional objectives within a single model. It provides flexible control over the strength of conditional guidance during inference, significantly advancing the development of text-to-X generation tasks~\citep{Rombach2022LDM,ho2022video,tevet2023human,liu2023audioldm}. The inpainting strategy~\citep{sohl2015deep} is applied for goal-conditioned trajectory generation by replacing the sampled values with the conditioning values after each denoising step~\citep{janner2022planning, huang2023diffusion}. Conditions can also be injected at all denoising steps for more stable conditioning~\citep{ni2023metadiffuser, chi2025diffusion}, typically through feature concatenation, Feature-wise Linear Modulation (FiLM)~\citep{perez2018film}, or cross-attention~\citep{vaswani2017attention} mechanisms.

\vspace{-4pt}
\section{Problem Definition}
\vspace{-2pt}
We consider the problem of generating trajectories given an initial configuration. 
Let $x_t\in\mathbb{R}^{3N}$ denote the 3D positions of $N$ particles at time $t$, and let $\alpha$ denote the associated physical attributes such as velocity, friction, mass, and elasticity: $\alpha = \{v, f, m, e\}$. We define the full physical state at time $t$ as $s_t = (x_t, \alpha)$, and the resulting trajectory as $s_{0:T} = \{s_0, s_1, \dots, s_T\}$.

To model the conditional distribution $p(s_{1:T}\mid s_0)$, Traditional approaches typically adopt an autoregressive formulation, sequentially predicting each subsequent state $s_{t+1}$ conditioned on the current state $s_t$. 
Instead, we propose an anchoring mechanism that conditions trajectory completion on the initial state ${s}_{0}$ and a predicted final state $\tilde{s}_{T}$, namely $p\bigl(s_{0:T} \mid s_0, \tilde{s}_{T}\bigr)$.

Given a dataset of particle-based physical trajectories $\mathcal{D}=\{s_0^{(i)}, \dots, s_T^{(i)}\}_{i=1}^N$, we aim to: (1) predict a plausible final state $\tilde{s}_{T}$ from $s_0$, accounting for uncertainty in attributes and interactions; and (2) complete the trajectory by generating $s_{1:T-1}$ conditioned on $(s_0,\tilde{s}_{T})$.

\vspace{-4pt}
\section{Method}
\vspace{-2pt}

The ReStep framework addresses the challenge of long-horizon physical dynamics prediction by reformulating trajectory generation as a future-anchored completion task. This reverse-consequence paradigm ensures that the global structure of the motion is constrained by a predictable future, effectively mitigating long-term drift.

This section details the ReStep architecture and is organized as follows: First, we introduce the \textit{Variational Steady-State Predictor} (Sec.~\ref{sec:stage1}), where a stochastic GNN-based VAE predicts a physically plausible final state $s_T$ conditioned on the initial state $s_0$ and physical parameters $\alpha$.
Second, we present \textit{Anchored Diffusion} (Sec.~\ref{sec:stage2}), which employs a Double-UNet architecture to synthesize a temporally coherent trajectory $s_{0:T}$ by interpolating between the two boundary anchor states.
Finally, we discuss ReStep-driven Gaussian rendering (Sec.~\ref{sec:stage3}), which integrates the predicted dynamics into a 3D Gaussian Splatting pipeline for physical inference and visual generation. 

\vspace{-4pt}
\subsection{Variational Steady-State Prediction}
\label{sec:stage1}
\vspace{-2pt}

To better capture the inherently stochastic nature of physical evolution, we adopt a VAE-style latent prediction architecture that models the multimodal uncertainty of object-level final states~\citep{vae}. In many object-interaction scenarios, small perturbations in the initial conditions (such as subtle variations in contact forces, collision angles, or surface friction) may lead to substantially different resting configurations. We encode the initial particle configuration into a latent distribution that explicitly represents the uncertainty over possible futures. 

Our method explicitly models particle-level interactions using Graph Neural Networks (GNNs)~\citep{sperduti1997supervised,hamilton2017inductive, gilmer2017neural}. Physical systems are inherently relational: the evolution of each particle is strongly influenced by its local neighborhood through contact, collision, and relative motion. GNNs naturally encode such relational inductive biases by performing message passing over dynamically constructed interaction graphs, enabling the model to reason about local geometric structure and inter-particle dependencies in a permutation-invariant manner.

Given an initial physical state $s_0 = (x_0, \alpha)$,  represented as a set of particles with positions and physical attributes, we first extract a global latent representation using a GNN-based encoder $\mathcal{E}_\theta(\cdot)$. The encoder constructs a $k$-nearest-neighbor graph in the particle space and performs multiple rounds of message passing conditioned on relative positions, aggregating local interactions into a global feature vector: $h_0 = \mathcal{E}_\theta(s_0)$.
The encoded representation $h_0$ is then passed to a learned prior network $\mathcal{P}_\theta(\cdot)$, which predicts the parameters of a latent Gaussian distribution: $[\mu_p, \log \sigma_p^2] = \mathcal{P}_\theta(h_0)$.

Meanwhile, during training, a posterior network $\mathcal{Q}_\phi(\cdot)$ is conditioned on the ground truth goal state $s_T$, which is first encoded using the same GNN-based encoder $h_T = \mathcal{E}_\theta(s_T)$. The posterior network then predicts the distribution parameters of the latent variable: $[\mu_q, \log \sigma_q^2] = \mathcal{Q}_\phi(h_T)$.

We sample $z_T\sim\mathcal{N}(\mu_q,\sigma_q^2)$, fuse it with the initial-state feature through a feedforward latent predictor
$\hat{h}_T=\mathcal{F}_\theta([h_0,z_T])$.
Finally, a GNN-based decoder $\mathcal{D}_\theta(\cdot)$ maps the predicted latent representation back to the particle space. Conditioned on both the global latent code $\hat{h}_T$ and the initial particle positions, the decoder performs relational message passing and predicts per-particle displacements, yielding the final predicted goal state:
$\hat{s}_T = \mathcal{D}_\theta(\hat{h}_T, s_0)$.

The model is trained using a composite objective, denoted by $\mathcal{L}_{\text{goal}}$, which combines a reconstruction loss on the predicted final state with KL divergence regularization that aligns the learned prior and posterior distributions:
\begin{equation}
\| \hat{s}_T - s_T \|_2
+ \beta \cdot D_{\text{KL}} \left(
\mathcal{N}(\mu_q, \sigma_q^2)
\;\|\;\mathcal{N}(\mu_p, \sigma_p^2)
\right).
\end{equation}
This GNN-based latent prediction framework enables the model to capture the multimodal nature of physical outcomes while explicitly accounting for inter-particle interactions and geometric structure. During inference, latent variables are sampled from the learned prior $\mathcal{P}_\theta(\cdot)$ to generate diverse and physically plausible final state hypotheses.

\begin{figure*}[t]
\vspace{-10pt}
    \centering
    \includegraphics[width=\linewidth]{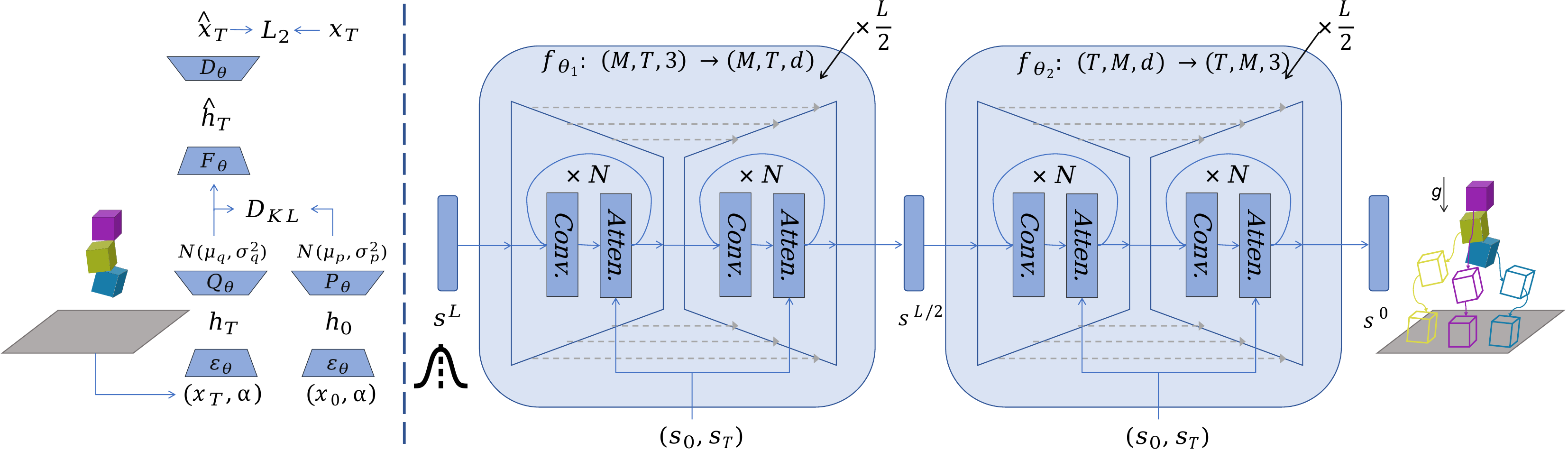}
    \vspace{-15pt}
    \caption{\textit{Model details.}
    \textbf{Left:} VAE-based steady-state predictor. Here, $x_T$ denote the positions at the final state, while $x_0$ and $\alpha$ represent the positions at the initial state and physical attributes. 
    \textbf{Right:} Denoising process. The diffusion-based UNet progressively refines the noisy trajectory $s^0$, where $s_0$ and $s_T$ correspond to the initial and final state conditions. $M$ denotes the number of particles and $T$ the length of the trajectory in time. This integration of initial and final state conditions enables trajectory completion.
    }
    \label{fig:stage1}  
\vspace{-5pt}
\end{figure*}

\vspace{-4pt}
\subsection{Anchored Diffusion}
\label{sec:stage2}
\vspace{-2pt}

We adopt a diffusion-based generative model to synthesize the entire trajectory $s_{0:T}$ given the initial state $s_0$ and the final state $s_T$. 
We will use $s$ to denote the entire trajectory $s_{0:T}$ for simplicity.


Notably, in particle-based physical systems, the number of particles $P$ typically varies with the number of objects or the resolution used to represent them. It is therefore desirable to train a model on varying particle counts, enabling it to capture a broader range of physical dynamics and generalize to novel scenes, such as those with different numbers of objects. We achieve this with the Double-UNet architecture, which applies convolutions separately along the time and particle dimensions. The first UNet models the temporal evolution of each particle, while the second captures the interactions among particles at each time step.


Formally, the Double-UNet is composed of multiple layers of convolution and attention operations. The convolution layers, denoted by $C(\cdot, \cdot)$, capture temporal evolutions or particle-wise interactions and aggregate information across diffusion steps. The attention layers, denoted by $A(\cdot, \cdot)$, perform cross-attention between the particle states and the conditional initial and final states to inject trajectory constraints, and self-attention among the particle states to capture local interactions and enforce physical consistency.
Let $h^{(i)}$ represent the feature at the $i$-th layer, which can be formulated as:
\begin{equation}
    h^{(i)} = C(h^{(i-1)}, l) + A(h^{(i-1)}, s_0, s_T), \quad i = 1:N,
\end{equation}
where $h^{(0)} = s^l$ is the noisy input trajectory, the diffusion step $l$ is incorporated by adding its sinusoidal positional embedding to $h^{(i-1)}$, and $N$ is the total number of layers. 

The denoising model is trained to minimize the following mean squared error (MSE) between the predicted clean samples and the ground-truth ones at uniformly sampled denoising steps, given by:
\begin{equation}
    \mathbb{E}_{l \sim \mathcal{U}\{1, 2, \dots, L\}, s \sim \mathcal{D}} \ \|s - f_{\theta}\left(\hat{s} \middle| s^l, l, s_0, s_T\right)\|^2,
\end{equation}
where $\mathcal{U}\{\cdot\}$ denotes the uniform distribution, $\mathcal{D}$ comprises trajectories with variable sequence lengths and particle counts, $s^l$ is obtained by corrupting the trajectory $s$ with Gaussian noise $\epsilon \sim \mathcal{N}(0, I)$, and $f_{\theta}$ denotes the denoising module that models the conditional distribution $p(s \mid s_0, s_T)$, allowing the generated trajectory to be anchored at both endpoints.






At inference time, we initialize from pure Gaussian noise and iteratively apply the learned reverse transitions to generate the full trajectory. This sampling process yields sequences that connect the initial state $s_0$ to the final state $s_T$.

\vspace{-4pt}
\subsection{ReStep-Driven Inverse Rendering}
\label{sec:stage3}
\vspace{-2pt}

While ReStep operates in particle space, the underlying particles and their physical parameters are often unobservable in real-world scenarios. To bridge this gap, we integrate ReStep as the physical engine within a 3D Gaussian Splatting (3DGS) pipeline. 
We then extend this framework to two tasks: (1) inferring physical parameters from visual observations, and (2) generating visual dynamics conditioned on initial frames and physical attributes.


\vspace{-4pt}
\subsubsection{Inverse Rendering for Physical Parameter Estimation}
\vspace{-2pt}

Given the initial and final multi-view images of a scene ($I_0,I_T$), we first reconstruct the 3D scene using a 3D Gaussian Splatting (3DGS) approach~\citep{kerbl20233d}. By extracting the centers of the optimized Gaussian primitives, we obtain object-level point clouds that capture the geometry at the beginning and the end of the motion.
These reconstructed point clouds are then mapped to the simulator particle representation to produce the initial and final particle states, $\hat{x}_0$ and $\hat{x}_T$. Specifically, we use the learned Gaussian segmentation, compute a bounding box for each object, and then sample simulator particles on and inside the box to approximate its spatial extent.
We denote $(\hat{x}_0, \hat{x}_T) \sim R(I_0,I_T)$, where $R(\cdot)$ is the 3DGS renderer. 
Based on these two states, we aim to recover the physical parameters that best explain the observed motion, including friction, restitution, and mass.

To achieve this, we leverage the pretrained steady-state predictor as a differentiable surrogate for forward physical prediction.
Since the module learns a mapping from the initial particle state and physical parameters to the final particle configuration, we can freeze the network and treat the physical parameters $\alpha$ as optimization variables.
The inference process is formulated as:
\begin{equation}
\alpha^{*} = \arg\min_{\alpha \in \Omega} \mathcal{L}_{\mathrm{inv}}\bigl(G_{\theta}\big(R(I_0), \alpha\big), R(I_T)\bigr),
\end{equation}
where $G_{\theta}$ denotes the pretrained steady-state predictor, $\Omega$ is the feasible parameter domain, and $\mathcal{L}_{\mathrm{inv}}$ measures the discrepancy between the predicted and observed final steady states. In practice, $\mathcal{L}_{\mathrm{inv}}$ can be defined as an $\ell_2$ loss.

\vspace{-4pt}
\subsubsection{ReStep-Driven Visual Dynamics Generation}
\vspace{-2pt}

Beyond inverse parameter estimation, our framework also supports visual dynamics generation. 
Given the initial images of a scene and its physical parameters (can also be estimated using our inverse rendering pipeline), we can simulate new visual dynamic trajectory. 
Similarly, we first reconstruct 3D scene and map its point clouds to the simulator particles, which are then fed into the ReStep dynamics model to predict future motions.
Specifically, we use the steady-state predictor to generate the final state under the given/estimated physical parameters, and use the anchored diffusion dynamics model to synthesize the full trajectory conditioned on the initial and final particle states. The particle trajectory is then converted into motion updates for the Gaussian scene representation. Additional details are provided in the Appendix~\ref{sec:rendering}.

Applying these updates to the reconstructed 3D scene yields dynamic point-cloud motion and, in turn, the corresponding visual dynamics.
Therefore, our method produces temporally coherent visual motion sequences that are consistent with both the observed initial appearance and physical properties. 
This enables controllable visual dynamics generation, where different physical parameters give rise to different predicted motions and visual outcomes.

\vspace{-6pt}
\section{Experiments}
\label{sec:experiments}
\vspace{-6pt}

\begin{figure*}[b]
\vspace{-5pt}
    \centering
    \begin{subfigure}[t]{0.325\textwidth}
        \centering
        \includegraphics[width=\linewidth]{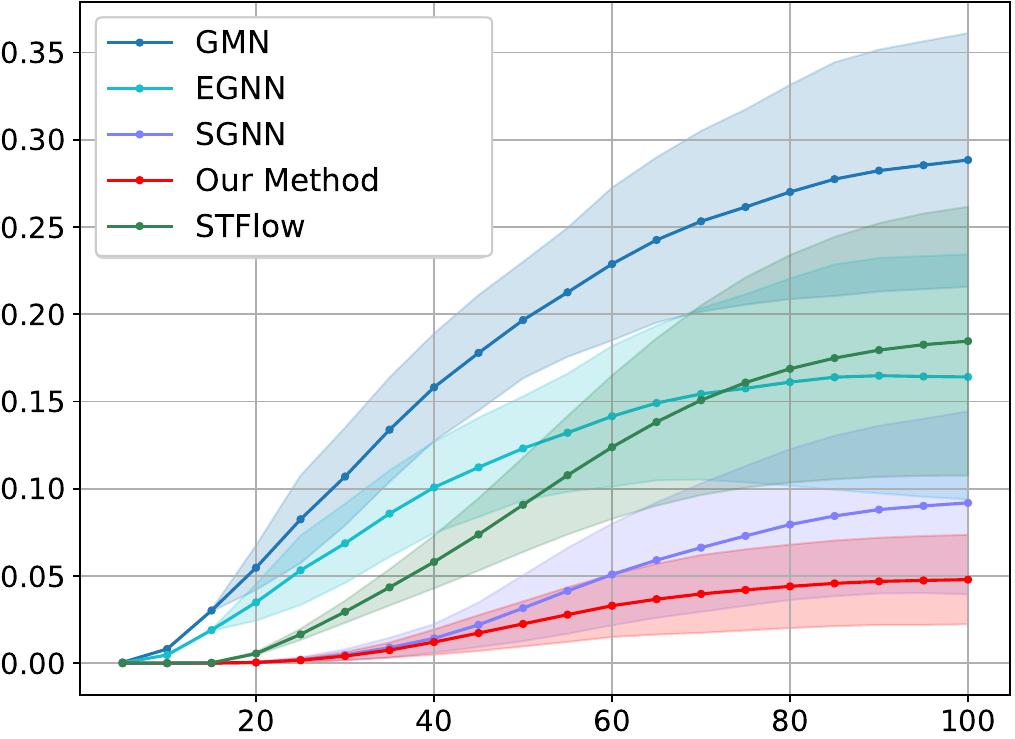}
        \caption{RigidFall}
        \label{fig:rigidfall_mse}
    \end{subfigure}
    \hfill
    \begin{subfigure}[t]{0.325\textwidth}
        \centering
        \includegraphics[width=\linewidth]{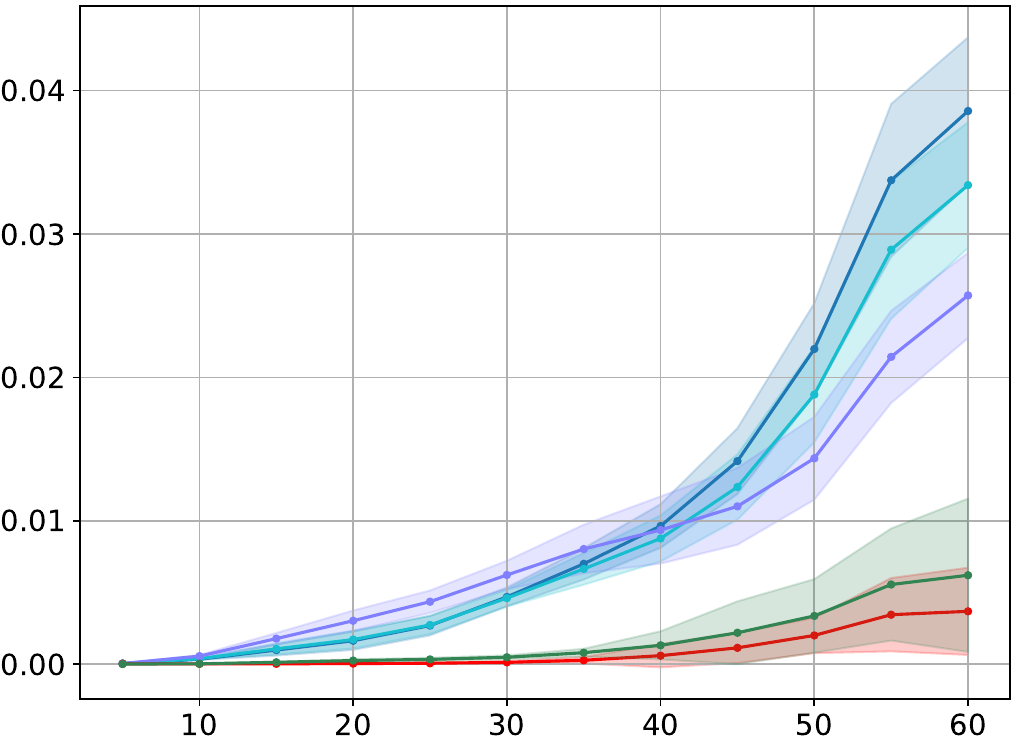}
        \caption{Topple}
        \label{fig:topple_mse}
    \end{subfigure}
    \hfill
    \begin{subfigure}[t]{0.325\textwidth}
        \centering
        \includegraphics[width=\linewidth]{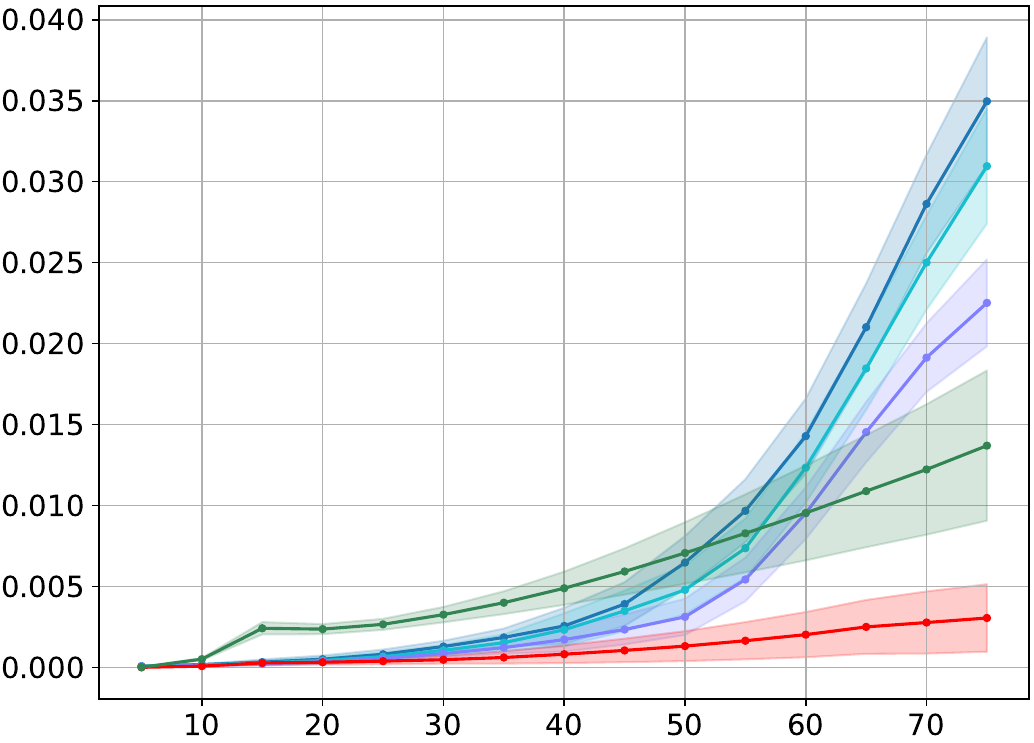}
        \caption{Drape}
        \label{fig:drape_mse}
    \end{subfigure}
    \vspace{-5pt}
    \caption{
    \textit{MSE comparisons across three FlexiFall scenarios between predicted and ground-truth trajectories.} 
    Our method consistently achieves the lowest error across all time steps.
    }
    \label{fig:traj_mse}
\end{figure*}

\begin{figure*}[t]
\vspace{-10pt}
    \centering
    \begin{subfigure}[t]{0.48\textwidth}
        \centering
        \includegraphics[width=\linewidth]{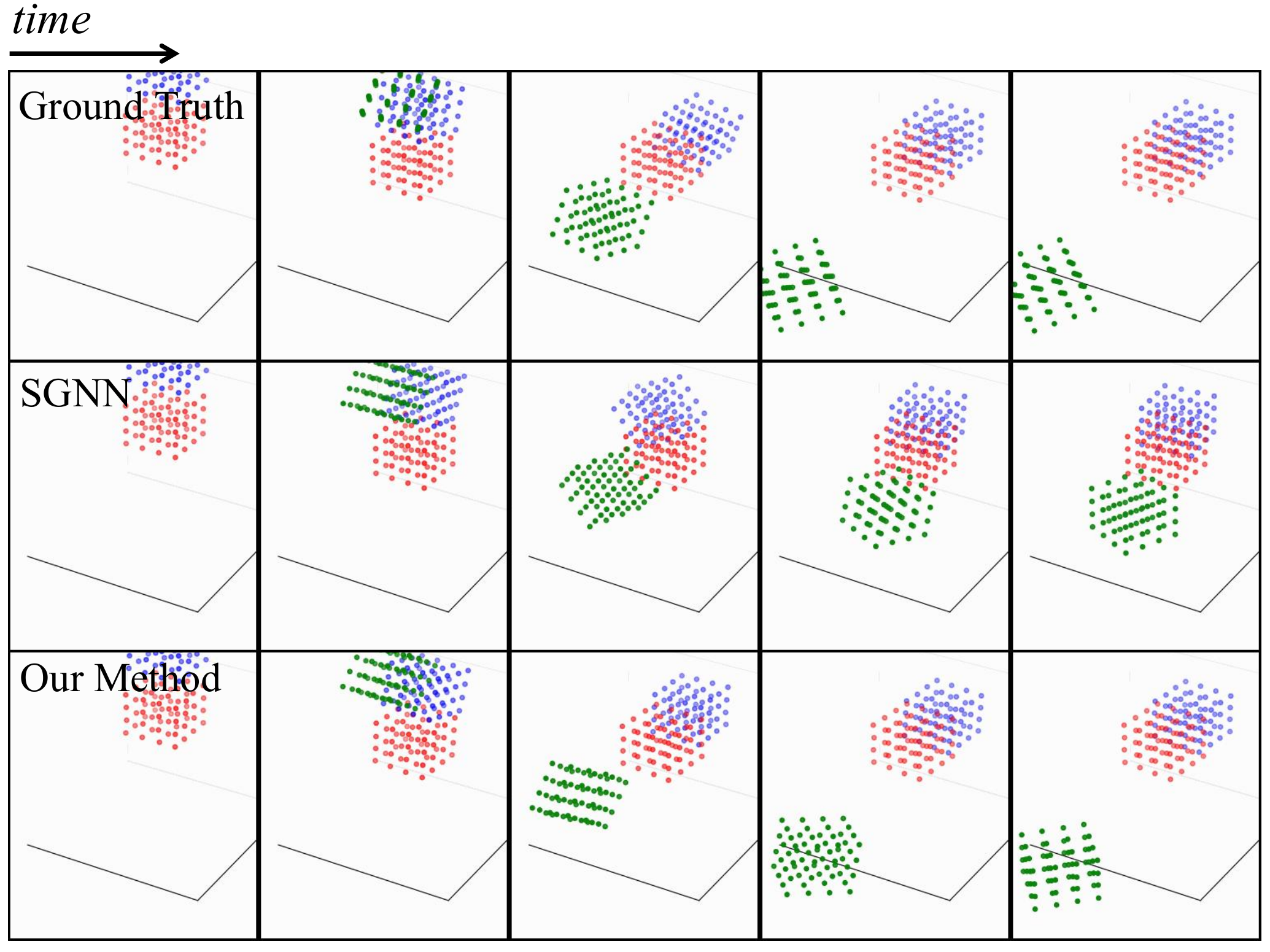}
        \label{fig:pointcloud_rigid}
    \end{subfigure}
    \hfill
    \begin{subfigure}[t]{0.48\textwidth}
        \centering
        \includegraphics[width=\linewidth]{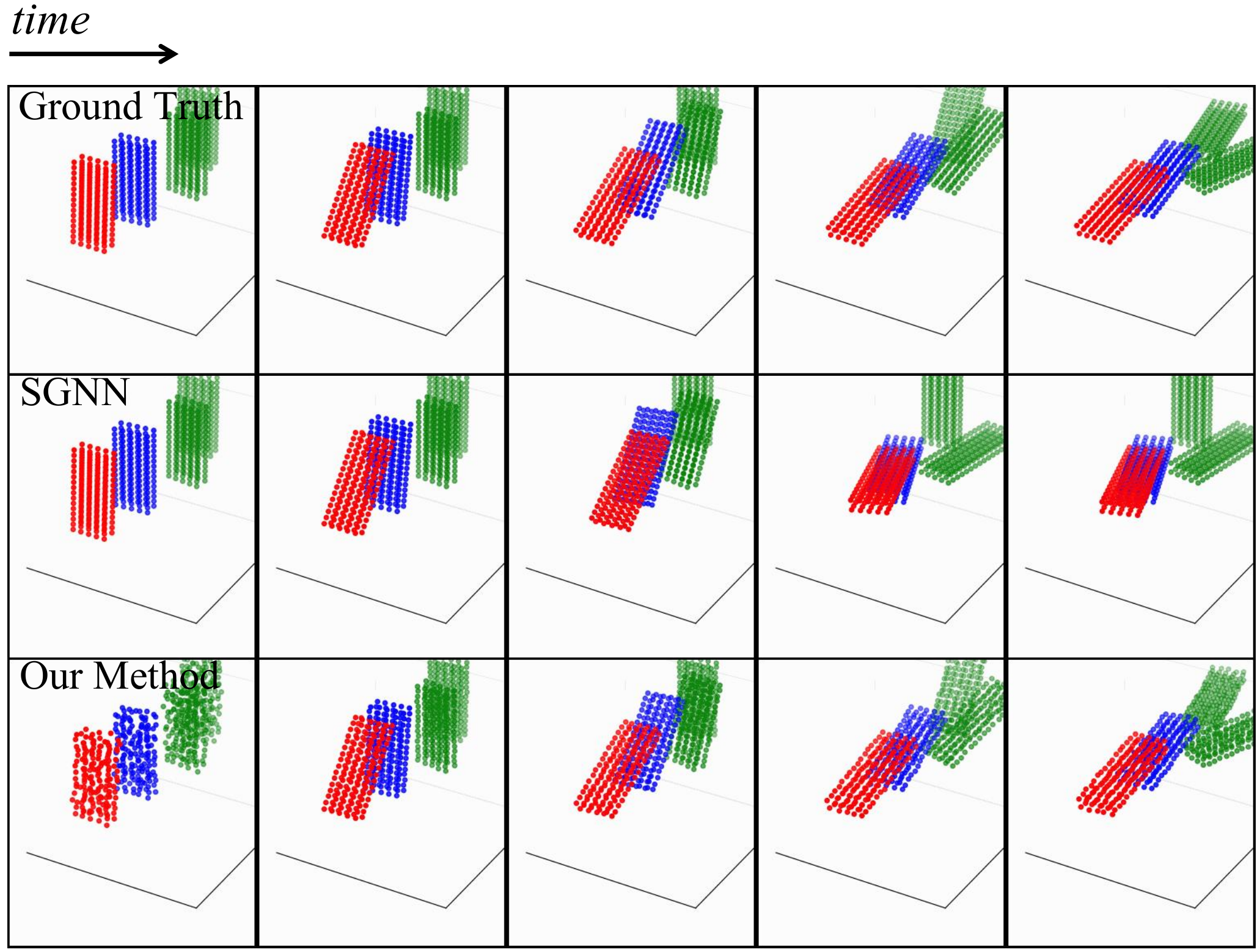}
        \label{fig:pointcloud_topple}
    \end{subfigure}
    \vspace{-10pt}
    \caption{\textit{Prediction showcases on the FlexiFall dataset}. Left: RigidFall scene. Right: Topple scene.}
    \label{fig:pointcloud_showcase}
\vspace{-5pt}
\end{figure*}

\subsection{Datasets}
\vspace{-2pt}

\begin{wraptable}{r}{0.45\columnwidth}
\vspace{-15pt}
\centering
\caption{\textit{Datasets.} We report dataset volume, horizon length, average number of particles, and average number of objects, respectively.
}
\vspace{-4pt}
\label{tab:dataset_statistics}
\small
\setlength{\tabcolsep}{4.5pt}
\begin{tabular}{lcccc}
\toprule
Scene & \#Seq & Horizon & \#Par & \#Obj \\
\midrule
\multicolumn{5}{c}{\textbf{FlexiFall}} \\
RigidFall & 10000 & 150 & 192 & 3 \\
Topple    & 10000 & 70--150 & 256 & 4 \\
Drape     & 5000  & 120 & 800 & 3 \\
\midrule
\multicolumn{5}{c}{\textbf{Physion}} \\
Collide   & 2150 & 151--452 & 2383 & 5 \\
Contain   & 2150 & 151--452 & 8666 & 7 \\
Drop      & 2150 & 151--302 & 3831 & 5 \\
Dominoes  & 2150 & 151--302 & 1909 & 7 \\
Roll      & 2150 & 151--302 & 3819 & 5 \\
\bottomrule
\end{tabular}
\vspace{-10pt}
\end{wraptable}
We evaluate our method on two datasets: \textit{FlexiFall} and \textit{Physion}. 
First, we construct \textit{FlexiFall} by extending \textit{RigidFall}~\citep{li2020visual} with additional scenes designed to diversify object interactions and deformation patterns. 
In addition to the original rigid cube falling setup, we introduce (i) a rigid \textit{Topple} scene featuring sequential contacts, and (ii) a cloth-draping scene in which a non-rigid sheet falls and settles onto rigid cubes. 
All scenes are simulated using \textit{Kubric}~\citep{greff2021kubric}. 
Second, we use \textit{Physion}~\citep{bearphysion}, which provides a wider spectrum of physical scenarios, including rigid and soft-body collisions, stable multi-object configurations, rolling, sliding, and projectile motion. We summarize the dataset statistics in Table~\ref{tab:dataset_statistics}.

\vspace{-4pt}
\subsection{Main Results on FlexiFall}
\vspace{-2pt}


During testing, given the initial state of a scene, the steady-state module performs a single random sampling to predict the final state. The predicted final state, together with the initial state, is then used as the condition for anchored diffusion to generate the full particle trajectory.

We first evaluate our method on the \textit{FlexiFall} dataset, including the original \textit{RigidFall} scene as well as the newly introduced \textit{Topple} and \textit{Drape} scenes. For quantitative evaluation, we compute the mean squared error (MSE) between the predicted and ground-truth particle trajectories, averaged every five time steps. In addition to autoregressive geometric baselines, we implement two recent generative trajectory baselines using the same particle-data interfaces: \textit{LatentFlow}, point-trajectory generation~\citep{boduljak2026what}, and \textit{STFlow}, spatio-temporal flow matching with a conditioned random-walk prior~\citep{tenbrinke2026stflow}. As shown in Figure~\ref{fig:traj_mse}, our method consistently achieves the lowest MSE across all time steps. Our method preserves long-term stability, resulting in trajectories that are more consistent with the ground-truth temporal evolution. Visualized predicted particle trajectories are shown in Figure~\ref{fig:pointcloud_showcase}.

\vspace{-4pt}
\subsection{Main Results on Physion}
\vspace{-2pt}

We evaluate our method on the \textit{Physion} dataset, which provides more complex physical scenarios. We follow the same quantitative evaluation protocol used on the previous dataset. As shown in Table~\ref{tab:physion_rollout_mse}, we report the rollout MSE of different methods at two prediction horizons, $t=50$ and $t=100$, and additionally evaluate LatentFlow with one generated trajectory per initial condition.

At the long-term horizon ($t=100$), our method consistently outperforms prior approaches across evaluated scenarios, demonstrating long-term prediction accuracy. This indicates that our model is more effective at mitigating error accumulation over long rollouts. 
At the shorter horizon ($t=50$), our method also achieves the best performance across evaluated scenarios, validating its strong short-term predictive capability. These results suggest that ReStep provides accurate and stable dynamics modeling across both short-term and long-term predictions.

\begin{table*}[t]
\centering
\caption{\textit{Rollout MSE on Physion for $t = 50$ and $t = 100$.} For each scenario, the best-performing method is highlighted in bold, while ‘–’ indicates that the method failed to produce a result at the corresponding horizon.}
\vspace{-4pt}
\scriptsize
\setlength{\tabcolsep}{4pt}
\label{tab:physion_rollout_mse}
\resizebox{\textwidth}{!}{%
\begin{tabular}{@{}lcccccccccc@{}}
\toprule
\multirow{2}{*}{{Method}} & \multicolumn{5}{c}{$t=50$} & \multicolumn{5}{c}{$t=100$} \\   
\cmidrule(lr){2-6} \cmidrule(lr){7-11}
& Collide & Contain & Drop & Dominoes & Roll & Collide & Contain & Drop & Dominoes & Roll \\ 
\midrule
EGNO~\citep{xu2024equivariant}      & 1.027& 0.111& 0.085& 0.011& 0.444& 1.703& 0.247& 0.107& 0.029& 0.604\\
GMN~\citep{gmn}       & 0.981& 0.131& 0.077& 0.012& 0.531& 1.592& 0.331& 0.120& --& 0.971\\
SGNN~\citep{sgnn}     & 1.009& 0.128& 0.066& 0.012   & 0.377& 1.365& 0.310& 0.091& 0.031& 0.616\\
LatentFlow~\citep{boduljak2026what} & 0.105& 0.308& 0.049& 0.005& 0.084& 0.261& 0.454& --& 0.014& 0.123\\
Ours      & \textbf{0.055}& \textbf{0.082}& \textbf{0.011}& \textbf{0.001}& \textbf{0.058}& \textbf{0.105}& \textbf{0.163}& \textbf{0.019}& \textbf{0.002}& \textbf{0.091}\\
\bottomrule
\end{tabular}
}
\vspace{-8pt}
\end{table*}

\begin{figure*}[b]
\vspace{-8pt}
    \centering
    \begin{subfigure}{0.49\linewidth}
        \centering
        \includegraphics[width=\linewidth]{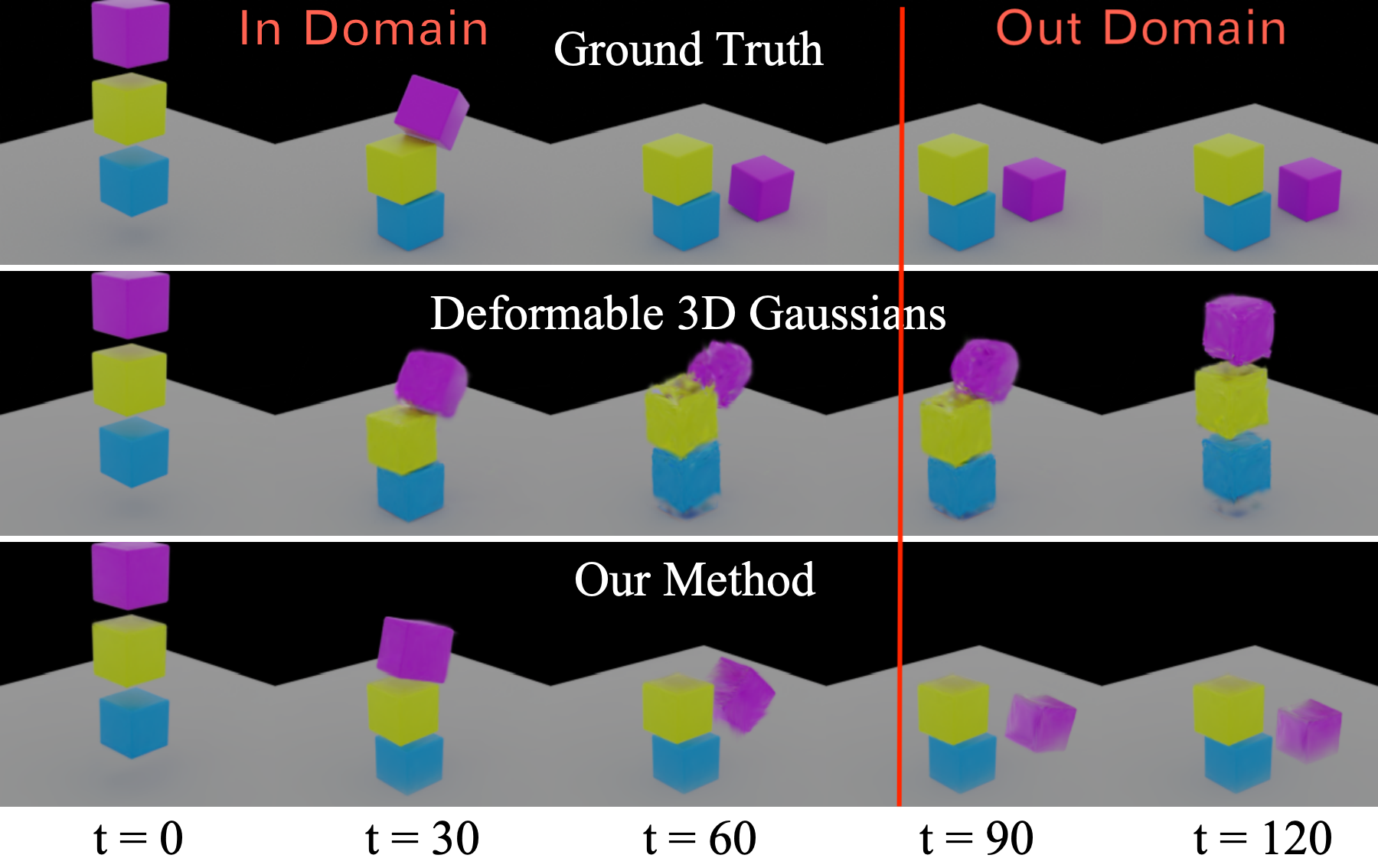}
        \caption{RigidFall}
        \label{fig:rigidfall}
    \end{subfigure}
    \hfill
    \begin{subfigure}{0.49\linewidth}
        \centering
        \includegraphics[width=\linewidth]{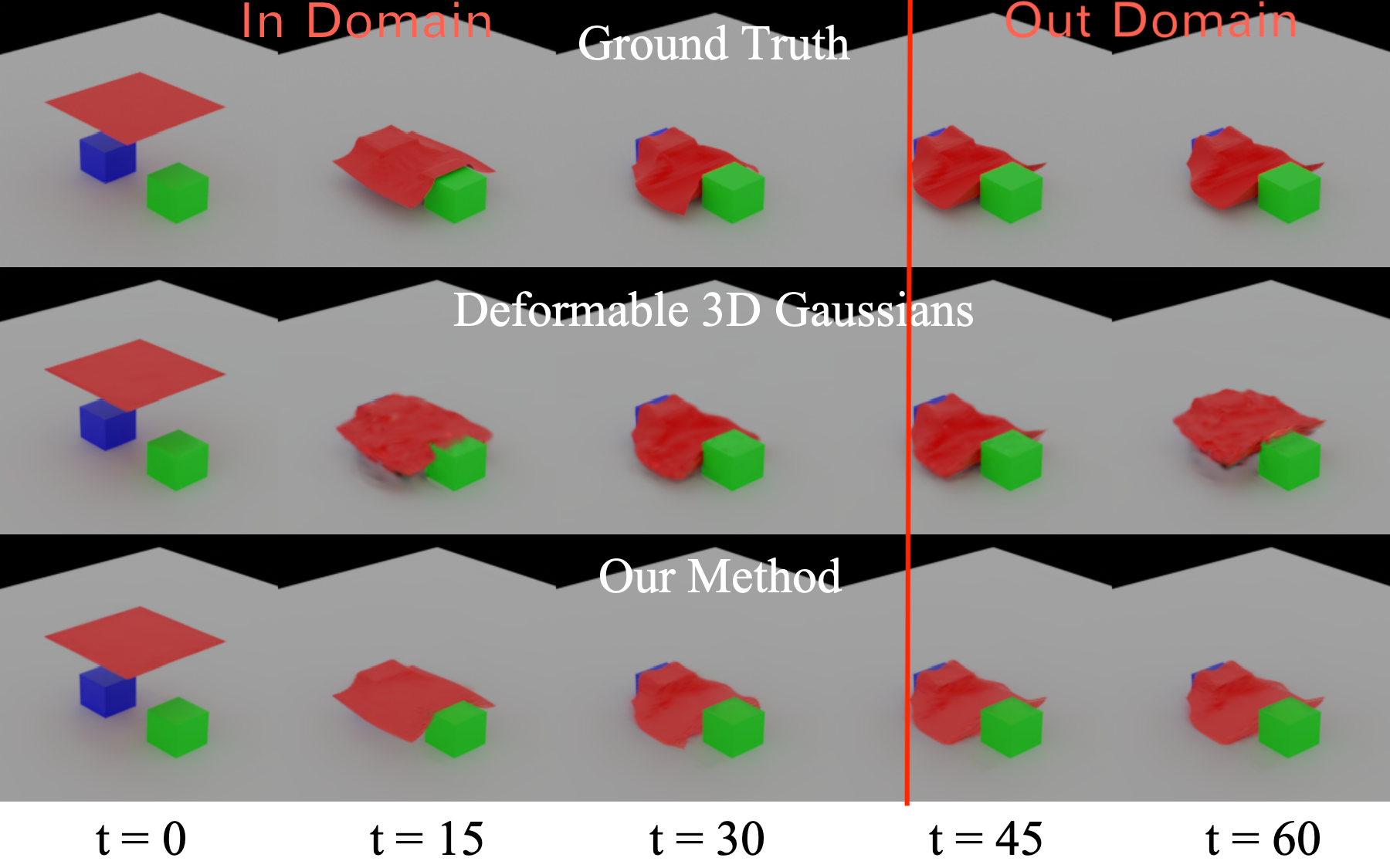}
        \caption{Drape}
        \label{fig:Drape}
    \end{subfigure}
    \vspace{-5pt}
    \caption{\textit{Comparison of rendered results.} Compared with Deformable 3DGS, our method illustrates that more accurate trajectory predictions for both rigid and non-rigid objects.
    }
    \label{fig:all_scenes}
\end{figure*}

\vspace{-4pt}
\subsection{Results on Physics-Grounded Rendering}
\vspace{-2pt}



\begin{wraptable}{r}{0.48\columnwidth}
\centering
\vspace{-10pt}
\caption{\textit{MSE of physical parameter estimation given visual observations.} We assess friction, restitution, and mass on the \textit{Topple} benchmark.}
\label{tab:inv_results}
\scriptsize
\setlength{\tabcolsep}{4pt}
\begin{tabular}{@{}lccc@{}}
\toprule
Method & Fric. & Rest. & Mass \\
\midrule
InvGraphics~\citep{jaques2019physics} & 0.1992 & 0.1998 & 0.1719 \\
Ours & \textbf{0.0889} & \textbf{0.0784} & \textbf{0.0796} \\
\bottomrule
\end{tabular}
\vspace{-6pt}
\end{wraptable}
We evaluate the capability of ReStep for system identification by performing physical parameter estimation on the \textit{Topple} scenario.
Leveraging the pretrained steady-state predictor as a differentiable physical surrogate, we invert for three key latent attributes, including friction, restitution, and mass, directly from visual observations.
Following the \textit{Kubric} convention~\citep{greff2021kubric}, these attributes are represented as scalar values normalized to the range $[0,1]$.
We construct a visual test set comprising paired scene observations and ground-truth physical parameters.
As shown in Table~\ref{tab:inv_results}, ReStep achieves a substantial reduction in estimation error across all parameters, outperforming the Physics-as-Inverse-Graphics baseline~\citep{jaques2019physics}.
Specifically, our method reduces the MSE from $0.1992$ to $\mathbf{0.0889}$ for friction, from $0.1998$ to $\mathbf{0.0784}$ for restitution, and from $0.1719$ to $\mathbf{0.0796}$ for mass.
These results demonstrate the effectiveness of our steady-state predictor, which not only improves forward dynamics but also provides more accurate and robust gradients for solving challenging inverse problems.

\begin{table*}[t]
    \centering
    \caption{\textit{Comparison of visual dynamics across three physical scenes.} 
    Metrics are reported for both \textbf{in-domain (seen during training)} and \textbf{out-domain (unseen future)} time steps. Our model achieves notably lower errors on out-of-distribution time steps, indicating stronger generalization.
    }
    \vspace{-4pt}
    \scriptsize
    \setlength{\tabcolsep}{5.5pt}
    \resizebox{\textwidth}{!}{%
    \begin{tabular}{@{}lccccccccc@{}}
    \toprule
   \multirow{2}{*}{{Method}} & \multicolumn{3}{c}{{Topple}} 
    & \multicolumn{3}{c}{{RigidFall}} 
    & \multicolumn{3}{c}{{Drape}} \\
    \cmidrule(lr){2-4} \cmidrule(lr){5-7} \cmidrule(lr){8-10}
     & SSIM↑ & PSNR↑ & LPIPS↓ 
                     & SSIM↑ & PSNR↑ & LPIPS↓ 
                     & SSIM↑ & PSNR↑ & LPIPS↓ \\
    \midrule
    \multicolumn{10}{c}{\textbf{In-Domain}} \\
    DeformGS~\citep{yang2024deformable}
         & 0.9846 & 43.2263 & 0.0156
         & 0.9784 & 35.1682 & 0.0227
         & 0.9859 & 41.3140 & 0.0164 \\
    Ours 
         & 0.9837 & 41.9273 & 0.0171
         & 0.9770 & 33.3704 & 0.0215
         & 0.9845 & 39.0993 & 0.0157 \\
    \midrule
    \multicolumn{10}{c}{\textbf{Out-of-Domain}} \\
    DeformGS~\citep{yang2024deformable}
         & 0.9662 & 27.4325 & 0.0551
         & 0.9707 & 29.0290 & 0.0335
         & 0.9796 & 35.2923 & 0.0226 \\
    Ours 
         & 0.9763 & 34.7931 & 0.0341
         & 0.9824 & 38.5410 & 0.0149
         & 0.9837 & 37.7353 & 0.0158 \\
    \bottomrule
    \end{tabular}
    }
    \label{tab:combined_inoutdomain}
\end{table*}


We further evaluate the visual scene generation capability of our framework on the \textit{FlexiFall} benchmark.
Starting from a short sequence of initial observations, we predict and render the remaining scene dynamics, and report quantitative results using PSNR, SSIM, and LPIPS. 
Our method is compared against the Deformable 3DGS baseline, with results summarized in Table~\ref{tab:combined_inoutdomain}.
Here, ``in-domain'' refers to time steps within the training distribution, while out-of-domain'' denotes long-horizon extrapolation to future time steps beyond the training horizon.
Notably, ReStep significantly outperforms Deformable 3DGS in the out-of-domain region. While purely appearance-driven models struggle with non-physical artifacts when extrapolating, our framework maintains high-fidelity structural consistency by leveraging future-anchored physical priors.
Qualitative results in Figure~\ref{fig:all_scenes} further illustrate that ReStep produces physically plausible trajectories that avoid the drift and distortions common in baseline methods.

\vspace{-4pt}
\subsection{Ablation Studies and Model Analyses}
\vspace{-2pt}

\begin{figure*}[t]
    \centering
    \begin{subfigure}{0.49\textwidth}
        \centering
        \includegraphics[width=\linewidth]{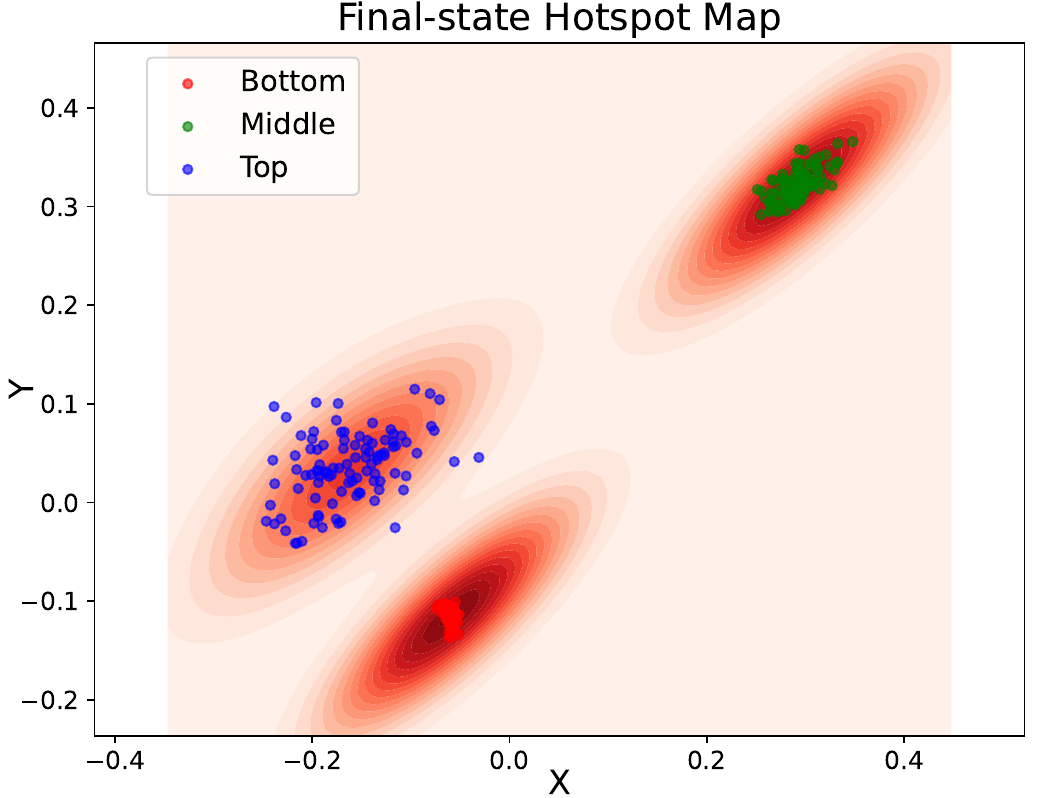}
        \caption{\textit{Heatmap of predicted steady-state distributions for three objects in RigidFall.} The highest object exhibits the widest spread and greatest uncertainty, while the lowest is more concentrated, reflecting the expressiveness of our model.}
        \label{fig:final_state_diversity}
    \end{subfigure}
    \hfill
    \begin{subfigure}{0.49\textwidth}
        \centering
        \includegraphics[width=0.95\linewidth]{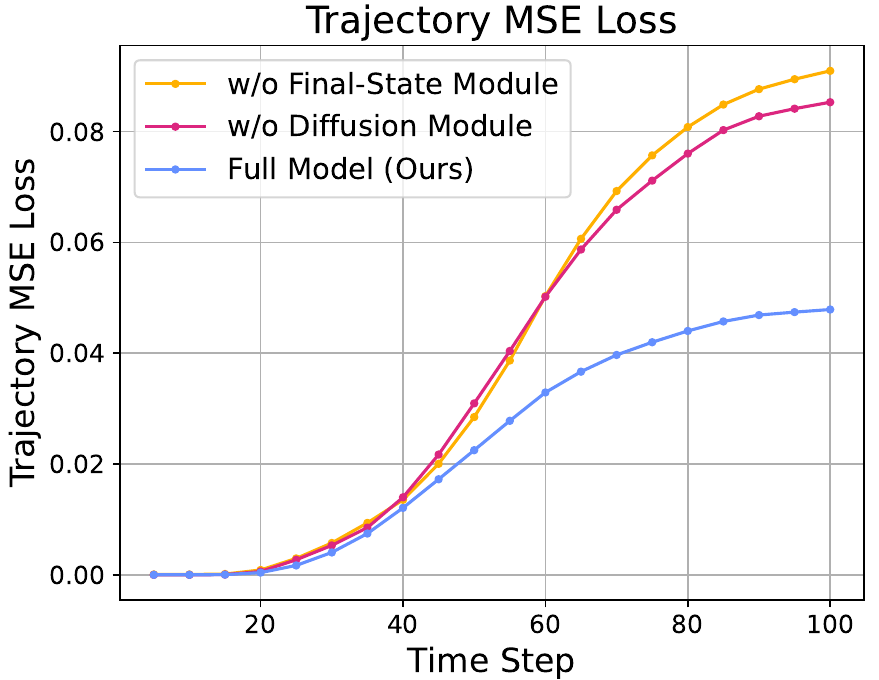}
        \caption{\textit{Ablation studies.} The full model (blue) maintains the lowest error, while removing the steady-state module (yellow) or the diffusion module (red) increases MSE, particularly at later steps. We use SGNN with steady-state inputs as the \textit{no-diffusion} baseline.}
        \label{fig:ablation}
    \end{subfigure}
    \vspace{-5pt}
    \caption{\textit{Prediction diversity of the steady-state predictor (a) and ablation results (b).}
    }
    \label{fig:heatmap_ablation}
\vspace{-10pt}
\end{figure*}

\subsubsection{Analysis of the Steady-State Predictor}
\vspace{-2pt}
To assess endpoint diversity, we sample $200$ trajectories in \textit{RigidFall} and plot each object's final XY-plane centroid (Figure~\ref{fig:final_state_diversity}). Objects starting higher exhibit broader endpoint distributions, whereas the lowest object is more concentrated, indicating that the predictor captures differences in outcome uncertainty.

We further assess the reliability of the predicted endpoint distribution using a calibration analysis. The empirical coverage is close to the nominal level at high confidence, while lower-confidence regions exhibit some under-coverage; the full procedure and results are reported in Appendix~\ref{sec:appendix_calibration}.

\subsubsection{Ablation Studies}
\vspace{-2pt}
Figure~\ref{fig:ablation} compares two variants on RigidFall: \textbf{(i) w/o steady-state prediction}, a non-autoregressive diffusion model conditioned only on the initial state; and \textbf{(ii) w/o anchored diffusion}, an autoregressive SGNN conditioned on the predicted steady state. Both variants increase long-horizon error, particularly the first, supporting the roles of future-state anchoring and diffusion-based synthesis. We further compare VAE-based and deterministic endpoint predictors on Contain and Drop, finding lower mean endpoint MSE with the VAE in both scenes (Appendix~\ref{sec:appendix_endpoint_stochasticity}). Moreover, Appendix~\ref{sec:additional_analysis} provides endpoint-condition, backbone, inference-cost, particle-count scaling, and failure-case analyses.

\vspace{-4pt}
\subsubsection{Out-of-Distribution Generalization}
\vspace{-2pt}
We evaluate the out-of-distribution (OOD) generalization under two settings: compositional scene shifts and temporal length variations.

\vspace{-8pt}
\paragraph{Compositional scene-composition shift.}
On Topple, we subsample objects from multi-object scenes to create unseen compositions while preserving the interaction rules. ReStep shows strong robustness: the steady-state MSE remains nearly unchanged from $0.0036\pm0.0020$ to $0.0035\pm0.0019$, while the trajectory-level MSE increases only moderately from $0.0010\pm0.0009$ to $0.0019\pm0.0013$.
This stability suggests that our GNN-based encoders effectively learn permutation-invariant relational priors that generalize across varying object counts.

\vspace{-8pt}
\paragraph{Variable trajectory lengths.}
On Dominoes, we assess generalization to shifts in temporal resolution.
We construct OOD samples by linearly interpolating trajectories to approximately twice their original length. Under this significant temporal shift, the average trajectory MSE increases only from $0.0035 \pm 0.0020$ (in-distribution) to $0.0055 \pm 0.0035$ (OOD).
This modest performance decay demonstrates that the Anchored Diffusion process is not overfitted to specific temporal scales but rather learns a robust, continuous mapping of physical evolution.

\vspace{-4pt}
\section{Conclusions and Limitations}
\vspace{-2pt}
In this work, we presented ReStep, an anchored framework for long-horizon dynamics prediction that conditions trajectory generation on both the initial and inferred final states. By first predicting a physically plausible final state and then generating intermediate states via an anchored diffusion model, ReStep effectively mitigates error accumulation and improves long-horizon stability compared to autoregressive methods. We further extended our framework to visually grounded tasks.

Despite these advantages, ReStep is still sensitive to errors in the predicted steady state, which can propagate through the diffusion model and produce physically implausible trajectories. This could be mitigated by rejection sampling or stronger consistency priors, which could motivate future work on more reliable final-state inference.

\subsection*{AI use statement}

We used generative AI tools to assist with LaTeX and citation formatting, language editing.
The authors reviewed the AI-assisted text. The authors take responsibility for the final content of this work.

\subsection*{Reproducibility Statement}
We report the evaluated datasets and scenes, trajectory horizons,
and evaluation metrics in Section~\ref{sec:experiments}.
The model architectures and implementation details are provided in
Appendix~C, while Appendix~D presents additional analyses of endpoint
conditioning, model backbones, inference cost, particle-count scaling,
and failure cases. We also provide our code in the supplementary materials. For the generative baseline comparison, Physion
LatentFlow results use one generated trajectory per initial condition;
the corresponding evaluation protocol and unavailable results are
identified in Table~\ref{tab:physion_rollout_mse}.

\bibliography{example_paper}
\bibliographystyle{iclr2027_conference}







\clearpage
\appendix

\section*{Appendix Contents}
\noindent
\hyperref[app:background]{A\quad Background}\dotfill
\pageref{app:background}\par
\hyperref[sec:rendering]{B\quad Details of ReStep-Driven Inverse Rendering}
\dotfill\pageref{sec:rendering}\par
\hyperref[app:architecture]{C\quad Architecture Details}\dotfill
\pageref{app:architecture}\par
\hyperref[sec:additional_analysis]{D\quad Additional Analyses and Experimental Diagnostics}
\dotfill\pageref{sec:additional_analysis}\par

\medskip

\section{Background}
\label{app:background}
\begin{table*}[!b]
\centering
\scriptsize
\caption{Network architectures and hyperparameters for Stage~1: Stochastic Goal State Prediction.
All MLP layers use ReLU activations unless otherwise specified.}
\label{tab:stage1_architecture}
\begin{tabular}{l|c|c|p{6.0cm}}
\toprule
\textbf{Module} & \textbf{Input Dim} & \textbf{Output Dim} & \textbf{Architecture Details} \\
\midrule

\textbf{Encoder} $\mathcal{E}_\theta(\cdot)$ 
& $N \times (3 + 3)$ 
& $z$
&
\begin{itemize}[leftmargin=10pt, topsep=0pt, partopsep=0pt, parsep=0pt, itemsep=1pt]
    \item Position encoder: MLP$(3 \rightarrow 128 \rightarrow 128)$
    \item Velocity encoder: MLP$(3 \rightarrow 128 \rightarrow 128)$
    \item Feature concatenation to form per-particle embeddings
    \item $L=3$ graph message-passing blocks with $k=16$ neighbors
    \item Linear projection: Linear$(256 \rightarrow z)$
\end{itemize}
\\
\midrule

\textbf{Posterior Network} $\mathcal{Q}_\theta(\cdot)$ 
& $z$
& $\mu_q,\,\log\sigma_q^2 \in \mathbb{R}^{z}$
&\begin{itemize}[leftmargin=10pt]
    \item MLP$(z \rightarrow 512 \rightarrow 2z)$ with normalization
    \item Output split into mean and log-variance
\end{itemize}
\\
\midrule

\textbf{Prior Network} $\mathcal{P}_\theta(\cdot)$ 
& $z$
& $\mu_p,\,\log\sigma_p^2 \in \mathbb{R}^{z}$
&\begin{itemize}[leftmargin=10pt, topsep=0pt, partopsep=0pt, parsep=0pt, itemsep=1pt]
    \item MLP$(z \rightarrow 512 \rightarrow 2z)$ with normalization
    \item Output split into mean and log-variance
\end{itemize}
\\
\midrule

\textbf{Frame Predictor} $\mathcal{F}_\theta(\cdot)$ 
& $z + z$
& $z$
& \begin{itemize}[leftmargin=10pt]
    \item MLP$(2z \rightarrow 512 \rightarrow z)$ with normalization
    \item Predicts the next latent particle state
\end{itemize}
\\
\midrule

\textbf{Decoder} $\mathcal{D}_\theta(\cdot)$ 
& $N \times (z + 3)$
& $N \times 3$
&
\begin{itemize}[leftmargin=10pt]
    \item Node encoder: MLP$((z+3) \rightarrow 256 \rightarrow 256)$
    \item $L=3$ graph message-passing blocks with $k=16$ neighbors
    \item Output MLP$(256 \rightarrow 256 \rightarrow 3)$
\end{itemize}
\\
\bottomrule
\end{tabular}
\end{table*}

\begin{table*}[!b]
\centering
\scriptsize
\caption{Network architectures and hyperparameters for Stage~2: Anchored Diffusion. 
We use the second UNet as an example, which applies convolution along the particle dimension.}
\label{tab:stage2_architecture}
\begin{tabular}{l|c|c|p{6.0cm}}
\toprule
\textbf{Module} & \textbf{Input Dim} & \textbf{Output Dim} & \textbf{Architecture Details} \\
\midrule

\textbf{Convolution Block} $C_\theta(\cdot,\cdot)$ 
& $(T, M, d_{in})$ 
& $(T, M, d_{out})$ 
&
\begin{itemize}[leftmargin=10pt]
    \item Conv1d$((T, M, d_{in}) \rightarrow (T, M, d_{out}))$, SiLU
    \item Feature fusion via element-wise addition
    \item Conv1d$((T, M, d_{out}) \rightarrow (T, M, d_{out}))$, SiLU
\end{itemize}
\\
\midrule

\textbf{Attention Block} $A_\theta(\cdot,\cdot)$ 
& $(T, M, d)$
& $(T, M, d)$
&
\begin{itemize}[leftmargin=10pt]
    \item SelfAttention$((T, M, d) \rightarrow (T, M, d))$
    \item CrossAttention$((T, M, d) \rightarrow (T, M, d))$
    \item GeGLU$((T, M, d) \rightarrow (T, M, 4d) \rightarrow (T, M, d))$
\end{itemize}
\\
\bottomrule
\end{tabular}
\end{table*}

\subsection{VAE-based Prediction}
Deep generative models have become a powerful framework for learning complex data distributions. 
Among them, the Variational Autoencoder (VAE)~\citep{vae} provides a probabilistic formulation that encodes data into a continuous latent space and reconstructs it through a decoder network. 
Formally, the encoder $q_\phi(z|x)$ maps an observation $x$ to a latent distribution, while the decoder $p_\theta(x|z)$ reconstructs $x$ from the latent variable $z$. 
The model is trained by maximizing the evidence lower bound (ELBO):
\begin{equation}
\mathcal{L}_{\mathrm{VAE}} = 
\mathbb{E}_{q_\phi(z|x)}[\log p_\theta(x|z)] 
- D_{\mathrm{KL}}(q_\phi(z|x) \| p(z)),
\end{equation}
which balances reconstruction accuracy and the regularization that enforces the latent space to follow a prior distribution.

Building upon this formulation, \textbf{VAE-based prediction} extends the generative modeling process to sequential or dynamic data. 
Instead of reconstructing the same input, the model learns to predict \emph{future states} by performing temporal modeling in the latent space. 
Specifically, given a latent encoding $z_t$ obtained from the current observation $x_t$, a transition model $p_\psi(z_{t+1}|z_t)$ predicts the next-step latent representation, which is then decoded to generate the predicted state $\hat{x}_{t+1}$. 
This structure enables the model to capture uncertainty, multi-modal dynamics, and long-term dependencies that are difficult to model directly in the observation space. 
Consequently, VAE-based prediction has been widely applied in video prediction, motion forecasting, and physical simulation tasks, where learning a compact and stochastic latent representation facilitates accurate and physically plausible future-state generation.

\subsection{Denoising Diffusion Probabilistic Models}

Denoising Diffusion Probabilistic Models (DDPMs) are a class of generative models that learn to reverse a fixed forward noising process. During training, Gaussian noise is progressively added to the data in $L$ steps, transforming structured samples into nearly pure noise. A neural network is then trained to de-noise these noisy samples step-by-step, learning to approximate the reverse of the diffusion process. At inference time, generation begins from random noise and proceeds iteratively using the learned reverse dynamics.

Formally, given a data distribution $q(x^0)$, the forward process adds Gaussian noise at each step:
\begin{equation}
    q(x^l \mid x^{l-1}) = \mathcal{N}(x^l; \sqrt{1 - \beta^l} x^{l-1}, \beta^l I),
\end{equation}
where $\beta^l$ is the noise schedule. The model is trained to approximate the reverse distribution $p_\theta(x^{l-1} \mid x^l)$ using a parameterized denoiser, often predicting either the original clean data $x^0$ or the added noise $\epsilon$.

Diffusion models offer strong sample diversity and high-fidelity generation, making them well-suited for tasks that require modeling complex multimodal distributions. In the context of physical dynamics, their iterative refinement process is particularly advantageous, as it allows for structured generation that respects spatial and temporal coherence. In our setting, we adapt diffusion models to synthesize physically plausible trajectories given an initial state and a predicted final state, leveraging their flexibility to model diverse motion sequences~\citep{ho2022video, blattmann2023stable}.

\section{Details of ReStep-Driven Inverse Rendering}
\label{sec:rendering}

In this section, we describe how diffusion-generated particle trajectories are converted into updates on the Gaussian scene representation. Since different objects exhibit fundamentally different physical behaviors, we separately handle non-rigid (cloth-like) objects and rigid bodies.

\subsection{Non-Rigid Objects}
For deformable objects such as cloth, global rigid transformations are insufficient to capture local stretching and bending. We therefore update each Gaussian point using a locally interpolated displacement derived from the diffusion particle trajectories.

Let $P^{(0)}$ and $P^{(t)}$ denote the initial and the $t$-th diffusion-generated particle positions for the cloth. For a Gaussian point $x$, we first identify its $k$ nearest cloth particles in the initial configuration. The displacement applied to $x$ is then computed as a weighted average of the corresponding particle displacements:
\[
\Delta x = \sum_{i=1}^{k} w_i \left(P_i^{(t)} - P_i^{(0)}\right), \qquad
w_i = \frac{\exp(-\|x-P_i^{(0)}\|)}{\sum_j \exp(-\|x-P_j^{(0)}\|)}.
\]
This formulation ensures that Gaussian points follow the local deformation field encoded by the diffusion particles while preserving smoothness and spatial coherence.  
Since non-rigid objects do not maintain a stable orientation, we only update point positions while keeping rotations unchanged.

\subsection{Rigid Objects}
Rigid bodies must preserve their shape, and therefore we determine a single global rigid transformation per object at each diffusion step. For an object, let $P^{(0)}$ and $P^{(t)}$ denote its initial and diffusion-predicted particle clouds. We compute the optimal rotation $R\in SO(3)$ and translation $t\in\mathbb{R}^3$ that align the initial cloud to the target cloud using the Kabsch algorithm:
\[
R, t = \arg\min_{R\in SO(3),\, t} 
\sum_i \big\| P_i^{(t)} - (R P_i^{(0)} + t) \big\|^2.
\]
The resulting rigid transform is applied to all Gaussian points belonging to the same object:
\[
\hat x = R x + t.
\]
This ensures shape preservation, consistent motion, and physically meaningful dynamics for rigid objects.

\subsection{Combined Update}
At each diffusion step, the Gaussian scene is updated by combining the non-rigid and rigid motions:
\[
X^{(t)} = X^{(t-1)} + \Delta X_{\mathrm{nonrigid}}^{(t)} + 
\Delta X_{\mathrm{rigid}}^{(t)}.
\]
This hybrid formulation allows the Gaussian scene to faithfully follow diverse physical behaviors, supporting accurate reconstruction of both deformable materials and solid objects.

\section{Architecture Details}
\label{app:architecture}
\subsection{Stage 1: Stochastic Goal State Prediction}

This section details the architectures of all sub-networks used in Stage~1.
The network descriptions below correspond to Table~\ref{tab:stage1_architecture},
which summarizes the full architectural specifications and hyperparameters.

Let $N$ denote the number of particles and $d=3$ the spatial state dimension.
Each particle is represented by its position and velocity.
All graph-based modules operate on per-particle features and model local interactions
using $k$-nearest-neighbor message passing.
Unless otherwise specified, all MLPs use ReLU activations.

\subsection{Stage 2: Anchored Diffusion}

Anchored Diffusion is implemented with the Double-UNet architecture composed of multiple layers of convolution and attention operations. The network descriptions below accompany Table~\ref{tab:stage2_architecture}, 
which summarizes the full hyperparameter specifications. Each convolution block has two 1D convolution layers as its trunk net, with SiLU as its activation function. The diffusion step embedding is added to its intermediate feature. Each attention block has one self-attention layer, one cross-attention layer for condition injection, and one GeGLU layer for element-wise gating. Residual connection and group normalization are adopted for both blocks for stable training. Downsampling is implemented using a convolution layer with strided kernels, and upsampling is implemented using nearest neighbor interpolation.

\section{Additional Analyses and Experimental Diagnostics}
\label{sec:additional_analysis}

\subsection{Calibration of the Endpoint Predictor}
\label{sec:appendix_calibration}
We assess whether the uncertainty produced by the steady-state predictor is calibrated by measuring whether the ground-truth final state falls within nominal credible regions of the predicted distribution. Table~\ref{tab:calibration} reports the nominal confidence level $p$, the empirical coverage, and the average empirical threshold, defined as the physical distance from the predicted mean required to cover the ground truth. Coverage is close to the nominal value at higher confidence levels: it reaches 0.8800 at $p=0.90$ and 0.9067 at $p=0.95$. At $p=0.50$, coverage is lower (0.3400), indicating under-coverage in more nonlinear contact-rich cases. The threshold increases monotonically with $p$, suggesting that the predicted uncertainty scales with the observed spatial variability.

\begin{table}[H]
\centering
\caption{Calibration of the steady-state predictor on the Dominoes scene.}
\label{tab:calibration}
\small
\begin{tabular}{ccc}
\toprule
Nominal confidence $p$ & Coverage & Threshold \\
\midrule
0.50 & 0.3400 & 0.004806 \\
0.80 & 0.7867 & 0.007224 \\
0.90 & 0.8800 & 0.009154 \\
0.95 & 0.9067 & 0.011467 \\
\bottomrule
\end{tabular}
\end{table}

\subsection{Sensitivity to the Endpoint Condition}
\label{sec:appendix_endpoint_conditions}
We compare three inputs to the trajectory generator: an endpoint predicted by the VAE, the observed endpoint (GT), and a perturbed endpoint (Random). The latter is constructed by adding clipped Gaussian perturbations to the initial condition. Table~\ref{tab:appendix_endpoint_conditions} reports trajectory MSE at two horizons. The GT condition is generally the most informative, whereas the predicted endpoint outperforms the perturbed control in all reported scenes. The predicted endpoint is not equivalent to the GT endpoint: the gap is particularly large on Roll at $t=100$, and the VAE condition slightly improves on GT for Drop at $t=50$. Thus, endpoint quality matters, but the observed ordering is not uniform across every scene and horizon.

\begin{table}[H]
\centering
\caption{Trajectory MSE under different endpoint conditions (lower is better). Random endpoints are constructed by adding clipped Gaussian perturbations to the initial condition. The VAE condition uses a stochastic endpoint sample.}
\label{tab:appendix_endpoint_conditions}
\begin{tabular}{llcc}
\toprule
Scene & Endpoint condition & $t=50$ & $t=100$ \\
\midrule
Dominoes & VAE & 0.0017 & 0.0026 \\
         & GT & \textbf{0.0012} & \textbf{0.0014} \\
         & Random & 0.0047 & 0.0073 \\
\midrule
Drop & VAE & \textbf{0.0110} & 0.0227 \\
     & GT & 0.0122 & \textbf{0.0174} \\
     & Random & 0.0400 & 0.0727 \\
\midrule
Roll & VAE & 0.0500 & 0.0753 \\
     & GT & \textbf{0.0206} & \textbf{0.0211} \\
     & Random & 0.3083 & 0.4796 \\
\bottomrule
\end{tabular}
\end{table}

\subsection{Endpoint-Predictor Backbone}
\label{sec:appendix_backbone}
Endpoint conditioning does not require a particular particle encoder. We replaced the GNN/$k$-NN endpoint predictor with a Transformer-style particle-set encoder and decoder, keeping the endpoint prediction objective. Table~\ref{tab:appendix_backbone} shows comparable or slightly lower mean endpoint MSE for the Transformer in these scenes. We therefore do not attribute the benefit of endpoint conditioning to the GNN backbone; the GNN remains a locality-based choice for modeling particle interactions.

\begin{table}[H]
\centering
\caption{Endpoint MSE (mean $\pm$ standard deviation; lower is better) for two endpoint-predictor backbones. These are endpoint prediction errors, not trajectory rollout errors.}
\label{tab:appendix_backbone}
\begin{tabular}{lcc}
\toprule
Scene & GNN/$k$-NN & Transformer \\
\midrule
Dominoes & $0.0100 \pm 0.0064$ & $0.0093 \pm 0.0074$ \\
Collide & $0.3137 \pm 1.2218$ & $0.2946 \pm 1.3407$ \\
Drop & $0.0357 \pm 0.0974$ & $0.0357 \pm 0.0928$ \\
\bottomrule
\end{tabular}
\end{table}

\subsection{Inference Cost and Particle-Count Scaling}
\label{sec:appendix_efficiency}
Table~\ref{tab:appendix_inference_cost} compares the reported inference costs of SGNN and the diffusion-based ReStep model. To compare a one-step autoregressive predictor with a full-trajectory generator, the reported FLOPs and time for ReStep are divided by trajectory length. On these two scenes, ReStep uses more FLOPs and parameters but has lower measured wall-clock time. This comparison reflects the evaluated implementations and hardware; FLOPs alone do not determine latency.

\begin{table}[H]
\centering
\caption{Reported normalized inference cost per predicted step. ReStep's full-trajectory FLOPs and time are divided by trajectory length; SGNN predicts one step at a time. The parameter count is that of the compared models.}
\label{tab:appendix_inference_cost}
\begin{tabular}{llccc}
\toprule
Scene & Model & FLOPs (B) & Time (s) & Parameters (M) \\
\midrule
Dominoes & SGNN & 63.254 & 1.258 & 1.551 \\
         & ReStep & 121.742 & 0.055 & 5.36 \\
Roll & SGNN & 8.727 & 1.016 & 0.993 \\
     & ReStep & 16.280 & 0.058 & 5.36 \\
\bottomrule
\end{tabular}
\end{table}

We separately profile the stage-1 GNN on scenes with different numbers of effective particles (Table~\ref{tab:appendix_particle_scaling}). Each forward pass builds a $k=16$ neighbor graph six times; static, non-informative large objects are excluded from the effective-particle counts. Within these evaluated scenes, total endpoint-predictor time ranges from 5.26 to 12.37\,ms as the particle count ranges from 474 to 2028. This stage-1 microbenchmark is distinct from the full-model comparison in Table~\ref{tab:appendix_inference_cost}, and should not be read as full trajectory-generation latency.

\begin{table}[H]
\centering
\caption{Stage-1 endpoint-predictor inference for different particle counts (batch size 1). Forward time excludes the separately measured $k$-NN construction; total time includes both.}
\label{tab:appendix_particle_scaling}
\begin{tabular}{lrrrr}
\toprule
Scene & Particles & FLOPs (B) & $k$-NN (ms) & Forward / total (ms) \\
\midrule
Collide & 361 & 14.81 & 1.53 & 4.05 / 5.58 \\
Roll & 474 & 19.45 & 1.49 & 3.77 / 5.26 \\
Contain & 603 & 24.75 & 1.52 & 4.02 / 5.53 \\
Drop & 1326 & 54.48 & 2.03 & 6.89 / 8.92 \\
Link & 1608 & 66.10 & 2.40 & 8.02 / 10.42 \\
Support & 2028 & 83.42 & 3.03 & 9.35 / 12.37 \\
\bottomrule
\end{tabular}
\end{table}

\subsection{Endpoint Errors and Failure Cases}
\label{sec:appendix_endpoint_failures}
Endpoint errors can propagate into the completed trajectory. We examine 100 RigidFall samples and divide trajectory MSE into quartiles. Of the 25 samples with the smallest endpoint errors, 23 also fall into the lowest trajectory-error quartile. Of the 25 with the largest endpoint errors, 21 fall into the highest trajectory-error quartile. All ten samples above the 90th percentile of endpoint MSE fall into the highest trajectory-error quartile; their mean endpoint MSE is 0.2688, mean trajectory MSE is 0.1277, and mean trajectory MSE at $t=100$ is 0.2488. These associations show that incorrect endpoints are a consequential failure mode. They do not, by themselves, establish how much of each trajectory error is caused by the endpoint error.

\begin{table}[H]
\centering
\caption{Association between endpoint-error groups and trajectory-error quartiles on 100 RigidFall samples. Rows count samples in the lowest and highest endpoint-error quartiles; the remaining 50 samples are omitted.}
\label{tab:appendix_endpoint_failures}
\begin{tabular}{lrrrr}
\toprule
Endpoint-error group & Traj. Q1 & Traj. Q2 & Traj. Q3 & Traj. Q4 \\
\midrule
Lowest 25\% ($n=25$) & 23 & 2 & 0 & 0 \\
Highest 25\% ($n=25$) & 0 & 0 & 4 & 21 \\
Highest 10\% ($n=10$) & 0 & 0 & 0 & 10 \\
\bottomrule
\end{tabular}
\end{table}

\subsection{VAE-Based versus Deterministic Endpoint Prediction}
\label{sec:appendix_endpoint_stochasticity}
To assess the benefit of the VAE-based endpoint predictor, we compare it with a deterministic endpoint predictor on the Contain and Drop scenes. As shown in Table~\ref{tab:appendix_endpoint_stochasticity}, the VAE achieves lower mean endpoint MSE in both scenes, reducing the error from 0.4052 to 0.3731 on Contain and from 0.0484 to 0.0357 on Drop, corresponding to relative reductions of 7.9\% and 26.2\%, respectively. These results provide empirical support for the VAE-based formulation in endpoint prediction.

\begin{table}[H]
\centering
\caption{Endpoint prediction errors for VAE-based and deterministic predictors on Contain and Drop. Lower mean endpoint MSE is highlighted in bold.}
\label{tab:appendix_endpoint_stochasticity}
\begin{tabular}{llc}
\toprule
Scene & Endpoint predictor & Endpoint MSE $\downarrow$ \\
\midrule
Contain & VAE                 & $\mathbf{0.3731} \pm 0.5481$ \\
        & Deterministic model & $0.4052 \pm 0.5824$ \\
\midrule
Drop    & VAE                 & $\mathbf{0.0357} \pm 0.0974$ \\
        & Deterministic model & $0.0484 \pm 0.1151$ \\
\bottomrule
\end{tabular}
\end{table}




\end{document}